\documentclass[letterpaper, 10 pt, conference]{ieeeconf}  %

\IEEEoverridecommandlockouts                              %

\usepackage{amsmath} %
\usepackage{amsfonts}
\usepackage{verbatim} %

\usepackage{mathtools}
\usepackage{booktabs}
\usepackage{lipsum}
\usepackage{float}
\usepackage{xspace}
\usepackage{cite}
\usepackage{graphicx}
\usepackage{wrapfig}
\graphicspath{{./images/}}
\usepackage{caption}
\usepackage{subcaption}
\usepackage[inline]{asymptote}
\usepackage{tikz}
\usepackage{hyperref}

\title{\LARGE \bf
ReF - Rotation Equivariant Features for Local Feature Matching
}

\author{Abhishek Peri$^{*1}$, Kinal Mehta$^{*1}$, Avneesh Mishra$^{1}$, Michael Milford$^{2}$,\\ Sourav Garg$^{2}$, K. Madhava Krishna$^{1}$ %
\thanks{*Denotes authors with equal contribution}%
\thanks{$^{1}$Robotics Research Center, IIIT Hyderabad.}%
\thanks{$^{2}$QUT Centre for Robotics, Queensland University of Technology (QUT), Australia.}%
\thanks{This research is supported by MathWorks.}
\thanks{\text{Code link: }\href{https://github.com/abhishek-peri/ReF-official-code}{https://github.com/abhishek-peri/ReF-official-code}}
}

\begin{document}

\makeatletter
\let\@oldmaketitle\@maketitle
\renewcommand{\@maketitle}{\@oldmaketitle
\centering
\includegraphics[width=0.9\textwidth]{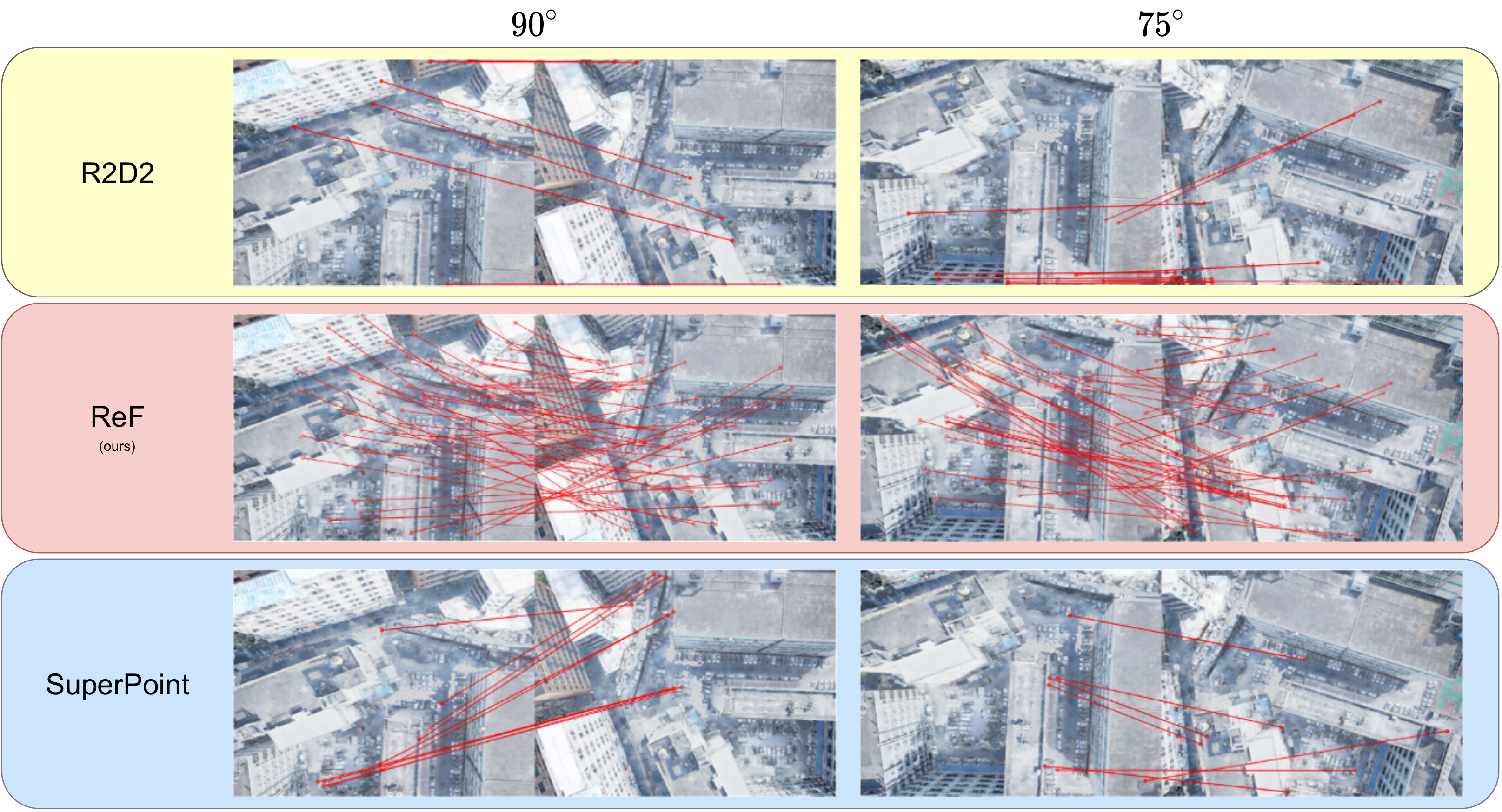}\\
\captionof{figure}{\small{ \textbf{Overview}.
\textit{Top} and \textit{Bottom}: R2D2 (made with standard CNN layers) and SuperPoint perform poorly where there are rotation viewpoint changes. 
\textit{Center}: ReF (Ours) provides more robust, dense, and correct correspondences under high viewpoint changes.
}
}

\label{fig:vpr_qualitative}
}

\makeatother

\maketitle
\thispagestyle{empty}
\pagestyle{empty}

\begin{abstract}
Sparse local feature matching is pivotal for many computer vision and robotics tasks. To improve their invariance to challenging appearance conditions and viewing angles, and hence their usefulness, existing learning-based methods have primarily focused on data augmentation-based training. In this work, we propose an alternative, complementary approach that centers on inducing bias in the model architecture itself to generate `rotation-specific' features using Steerable E2-CNNs, that are then group-pooled to achieve rotation-invariant local features. We demonstrate that this high performance, rotation-specific coverage from the steerable CNNs can be expanded to all rotation angles by combining it with augmentation-trained standard CNNs which have broader coverage but are often inaccurate, thus creating a state-of-the-art rotation-robust local feature matcher. We benchmark our proposed methods against existing techniques on HPatches and a newly proposed UrbanScenes3D-Air dataset for visual place recognition. Furthermore, we present a detailed analysis of the performance effects of ensembling, robust estimation, network architecture variations, and the use of rotation priors. 

\end{abstract}

\section{INTRODUCTION}

Local feature matching plays a central role in many computer vision and robotics tasks such as visual place recognition~\cite{cummins2011appearance,hausler2021patch,garg2021your}, image stitching~\cite{image_stitching}, and 6-DoF pose estimation~\cite{sattler2018benchmarking,superglue,rord,d2net}. 
Traditional approaches for detection and description such as SIFT~\cite{sift} and ORB~\cite{orb} first detect the points of interest and then use the patch around the pixel/keypoint to generate a descriptor. These approaches fail under severe illumination and rotation changes. Recent learning-based approaches such as SuperPoint~\cite{superpoint}, SuperGlue~\cite{superglue}, and RoRD~\cite{rord} are trained as a `general' system for handling such challenges using training data that is augmented heavily with illumination and viewpoint variations.

In this work, we take an alternative but a complementary approach that induces bias in the model architecture itself to generate `rotation-specific' features. For this purpose, we make use of Steerable CNN~\cite{scnn, e2cnn} and propose ReF (Rotation Equivariant Features) network which generates feature descriptors that are group pooled to achieve rotation-invariant local feature matching. We then show that highly accurate but rotation-specific coverage of ReF can be expanded to all rotation angles by combining it with data augmentation-trained RoRD~\cite{rord} which has better rotation coverage but is often inaccurate, thus leading to an overall state-of-the-art rotation-robust local feature matching. We make the following contributions:
\begin{enumerate}
    \item We introduce novel Rotation-equivariant Features, dubbed ReF, based on steerable E2-CNNs which can be group-pooled to achieve rotation-invariance for local feature matching, thus leading to $0^\circ$-like matching performance under different rotations including $90^\circ$ and $180^\circ$.
    \item We present a `correspondence ensemble' of E2-CNN based ReF with data-augmented standard CNN based RoRD~\cite{rord}, improving the overall coverage of performance response to different rotation angles.
    \item We introduce ``UrbanScenes3D-Air'', a new dataset that enables benchmarking Visual Place Recognition from a drone's view under varying directions of travel. 
    \item While evaluating local feature matching both standalone and on the VPR task, we present a detailed analysis of steerable CNNs for rotation-robustness under the effect of ensembling, robust estimation (RANSAC), network architecture variations, and the use of rotation priors.
\end{enumerate}

\section{RELATED WORK}
Here we briefly discuss local feature matching and rotation equivariant networks.

\subsection{Local Feature Matching}
Local feature matching plays a central role in many computer vision applications such as visual place recognition, image stitching, pose estimation, etc. Traditional approaches like SIFT~\cite{sift}, ORB~\cite{orb} try to address specific problems like scale invariance, or rotation invariance in the descriptors. But their performance heavily depends on the rules used for keypoint detection and description. 

Recent works leverage CNNs for keypoint detection and description. D2-Net~\cite{d2net} uses detect and describe approach which is jointly trained using triplet loss. SuperPoint~\cite{superpoint} uses a self-supervised approach to learn the keypoints and descriptors. R2D2~\cite{r2d2} uses prediction of reliability map and repeatability map to find unique features which are repeatable under some transformations. These approaches perform well under challenging conditions where standard rule based systems fail.

Some works for learning rotation invariant descriptors include LBP-HF~\cite{LHF, lbp}, which leverages local binary pattern histograms to learn invariant features. RI-LBD~\cite{lribp} uses learning based system to learn invariant local descriptors. RoRD~\cite{rord} uses data augmentation to learn rotation robust descriptors. These approaches try to obtain descriptors that are rotation invariant. Using augmentation during training does improve the performance compared to standard methodologies~\cite{rord}.
Our approach, on the other hand, tries to make the model inherently rotation equivariant. This provides a model which can generalize better over high rotation changes. 

\paragraph*{Visual Place Recognition (VPR)} VPR is commonly used to describe the ability to recognize a previously visited location despite significant appearance or viewpoint changes~\cite{garg2021your}. Local feature matching is also a significant part of VPR pipelines~\cite{cummins2011appearance}.  Recent works have leveraged advances in deep networks based patch aggregation~\cite{hausler2021patch}, saliency detection~\cite{keetha2021hierarchical}, and transformers with multi-scale attention~\cite{wang2022transvpr}. As per our knowledge, our work is the first to utilize a rotation equivariant system for the task of VPR and demonstrate state-of-the-art results on real-world scenes from our \textit{UrbanScenes3D-Air} dataset.

\subsection{Rotation Equivariance Networks}
Equivariance is a property where a transformation in the input space leads to the corresponding transformation change in the feature space. 
CNNs are known to be translation equivariant, which means that a translation change in input space leads to a corresponding translation change in feature space. Rotation equivariance is a property desired by a lot of vision applications in medical images, robotics for drones, etc.
Steerable CNN is the tool used to achieve this property \cite{scnn, e2cnn}. The idea is to learn group representations that provide equivariance under the actions of the elements of the group. Jenner provides a framework unifying such convolutions and differential operators in \cite{scnn_pdo}.
Cesa and Weiler derive a specific implementation of these concepts for the euclidean group $E(2)$ (group of isometries in $E^2$ space - rotation, translation, mirroring) in~\cite{e2cnn}.

The efficacy of equivariant deep learning has been explored in the field of medical imaging~\cite{mri_roteq}, object detection~\cite{redet}, aircraft detection~\cite{aircraft_detection} and reinforcement learning~\cite{rl_geq}. To the best of our knowledge, this concept has not been applied to feature description to obtain descriptors that are rotation equivariant.
We demonstrate the use of this concept in local feature matching by designing a novel architecture that is trained using concepts inspired from~\cite{r2d2}.
Our architecture increases the descriptor matching performance significantly over a wide range of rotational viewpoint changes.

\begin{figure*}[t]
    \centering
    \vspace*{0.2cm}
    \includegraphics[width=\textwidth]{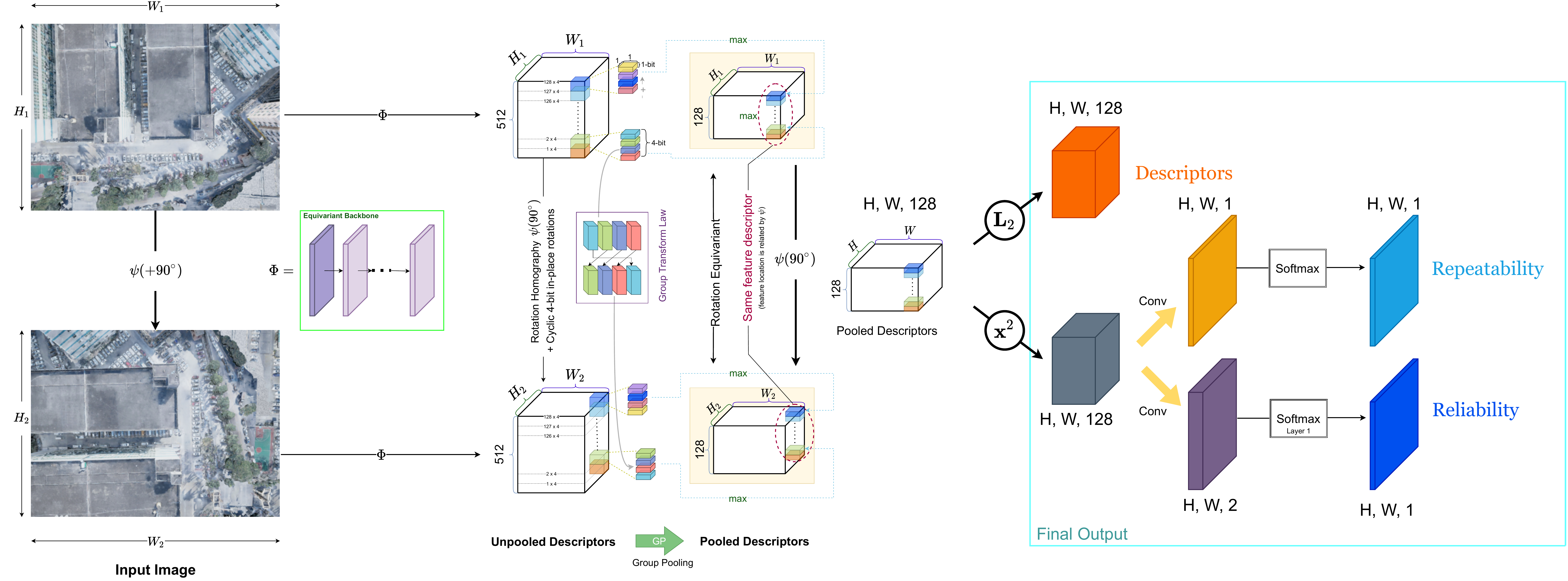}
    \caption{Model and Steerable CNN Implementation - $C_4$ equivariant ($n=4$).
        \textbf{Left}: The input is presented as a \emph{trivial representation} (RGB image), and passed through the rotation equivariant layers $\Phi$.
        \textbf{Center} \emph{unpooled} descriptors: The result of the convolution follows the regular representation (n-bit in-place rotation along the descriptor channel).
        \textbf{Center} \emph{pooled} descriptors: The group pooling is the \texttt{max} substitution of n-bits in the descriptor (of each feature). Note that the pooled descriptors of features in images related by pure rotation are \emph{equal} (therefore rotation equivariant). The features themselves are rotation invariant (they do not change based on rotation).
        \textbf{Right} (output): The feature \emph{descriptors} are obtained by taking the $L_2$ norm of the pooled descriptors. The \emph{repeatability} and \emph{reliability} maps (keypoints) are obtained through a standard CNN and a softmax.
        }
    \label{fig:steer_eq_gp}
    \label{fig:model_arch}
\end{figure*}

\section{METHODOLOGY}
This paper aims to develop rotation invariant descriptors for local feature matching by inducing bias in the model architecture itself, complementary to learning invariance through data augmentation based training. We achieve this by using Steerable CNNs instead of standard/conventional CNNs. In this section, we first introduce the Steerable CNNs, then describe our specific model architecture and its training methodology, which is then followed by local feature matching and ensemble strategy.

\subsection{Steerable CNN}
\label{sec:scnn}

Standard CNNs operating on a 2D image grid are translation equivariant. Steerable CNNs enable us to extend the property of equivariance to groups other than $(\mathbb{R}, +)$ (action group of 2D translations) using group theory. Cohen et al. introduced Steerable CNNs in~\cite{scnn}. For an input image $x$, an \textit{action group} such as a rotation homography $H$ can be applied such that the output of the function $f$ is steerable (that is, \emph{H-equivariant}) if the criterion in the equation~\ref{eq:steer-cnn} is met.

\begin{equation}
    \left(f \circ \pi_i(h) \right )(x) = \left ( \pi_o(h) \circ f \right ) (x)
    \quad
    h \in H
    \label{eq:steer-cnn}
\end{equation}
where the functions $\pi_i$ and $\pi_o$ are the input and output group representations. As described in~\cite{lin-rep-fg-book}, linear representations of finite groups allow representing group actions as linear morphism, where rotation can be implemented as a homography group representation. Specifically, we employ the action group $H = C_n$ which consists of rotations $\{ \theta_p \mid \theta_p = \frac{2\pi}{n}p \,,\;\;\; p \in \mathbb{Z}/n\mathbb{Z}\}$. The group actions are defined on a particular \emph{field type}. 

The raw RGB image input to our network uses a `trivial representation' of the $C_n$ group as its field type, which
applies no transformation (identity map) and presents images as RGB images. The field type of output and intermediate layers uses the `regular representation', which cyclically rotates the elements, as shown under \texttt{Group Transform Law} in the model architecture (in Figure~\ref{fig:model_arch}). In this form, we refer to the output descriptors as the \emph{unpooled} descriptors (the $n$ \emph{bits} in descriptors rotate - the representation for $n=4$ is shown in Figure~\ref{fig:steer_eq_gp}). Group pooling, which is in-place substitution by the maximum of the n-bits, converts these unpooled descriptors to \emph{pooled} descriptors. This, as shown in Figure~\ref{fig:steer_eq_gp} (center region - from unpooled to pooled), makes the descriptors rotation equivariant where the rotated versions (of feature descriptors) are only related by a rotation homography (highlighted in the pooled descriptors in center of Figure~\ref{fig:steer_eq_gp}). Note that the pooled descriptors of specific pixels have now become invariant to rotation by the angles in $C_n$ (the vector doesn't change, only its location in the output space changes).

\subsection{Model Architecture}
\label{subsec:model-arch}
Our architecture, shown in Figure~\ref{fig:model_arch}, consists of 5 layers of E2-CNN with increasing channels: (32, 64, 128, 256, 512). Each E2-CNN layer is followed by a batch normalization layer and ReLU non-linearity.
These layers are followed by a group-pooling layer (see the center of Figure~\ref{fig:model_arch}) which pools the feature field across different rotations to get equivariant features. That is, for each input field, an output field is built by taking the maximum activation within that field; as a result, the output field transforms according to a trivial representation. This is required as the computations of standard CNN needs trivial representation. Moreover, the feature map obtained after group pooling is equivariant under rotation as well as translation, this is because E2-CNNs also strides along the x-axis and y-axis as the standard CNNs.
We implement the network for $C_8$ space, therefore group pooling reduces the number of channels from 512 to 64.
These pooled features are then passed through a standard CNN to get a reliability map. Similar operation is done to get the repeatability map by passing the pooled features through a standard CNN. Softmax is applied on the reliability map and softplus is applied on the repeatability map as non-activation. The pooled features are $L2$-normalized to get the pixel level descriptors that can be used for feature matching.
The model is trained using training methodology of R2D2~\cite{r2d2}. We use R2D2 as the base training methodology as it enables us to train for feature detection and description from scratch. 
Repeatability map is trained jointly using two losses, one which tries to maximize the cosine similarity between the repeatability maps of two images, and the other which tries to maximize the local peakiness of the repeatability map.
Reliability map is basically the discriminative score of the descriptors. Descriptor matching can be seen as a ranking optimization problem and hence tries to maximize for differentiable average-precision defined in~\cite{aploss}.

\subsection{Correspondence Ensemble}
We follow similar ensemble strategy as mentioned in~\cite{rord}, though we use two separate networks to generate the keypoints, descriptors, and matches.
Our ensemble strategy is based on extracting the top correspondences from the pool based on the cosine similarity between their descriptors. Once we have keypoints and descriptors from our model ReF and the RoRD model~\cite{rord}, mutual nearest-neighbor based feature correspondences are obtained between an image pair independently for each model. Using the descriptor similarity of the combined set of correspondences obtained from the two models, we select the top 50\% of the correspondences which have the highest cosine similarity values. These correspondences are then filtered by RANSAC-based geometric verification to obtain a final set of feature correspondences. In Section~\ref{sec:results_hpatches}, we present results for both the settings, that is, without RANSAC (using only cosine similarity based ensemble) and with RANSAC. 

\subsection{Feature Matching}
We use mutual nearest neighbor for finding the correspondences between the image pair once we have the keypoints and corresponding descriptors. We apply RANSAC on the obtained matches to filter out noisy matches. We can see in section \ref{sec:results_hpatches} that our model is able to perform well on different rotation variations. RANSAC helps us highlight the fact that our model indeed learns correct correspondences.

\begin{figure}
    \centering
    \vspace*{0.2cm}
    \includegraphics[width=0.47\textwidth]{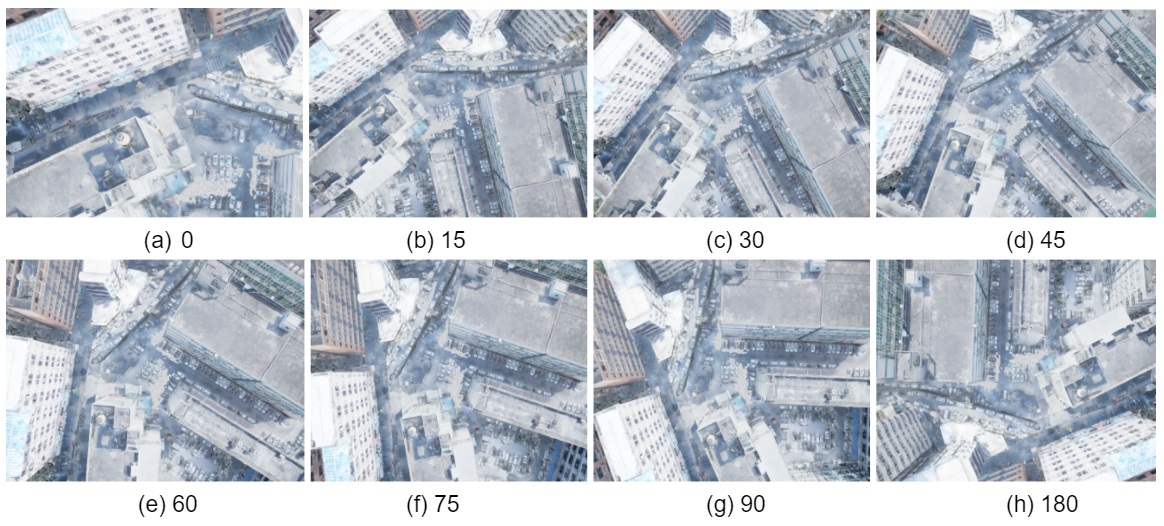}
    \includegraphics[width=0.25\textwidth, height = 0.25\textwidth]{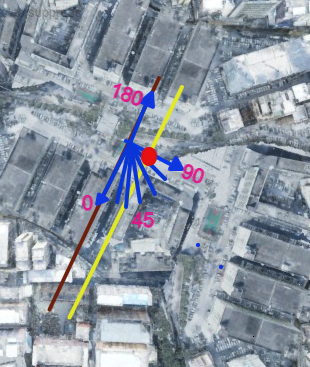}
    \includegraphics[scale=1]{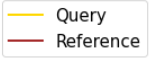}
    \caption{The drone traverses along two trajectories of which the yellow corresponds to the query and the brown correspond to the reference. At a given location, 8 different images are recorded by the drone during its time of flight in the brown trajectory. The sample images corresponding to the 8 cameras can be seen here which have a relative degree of rotation with respect to the yaw axis of the drone.}
    \label{fig:airsim}
\end{figure}

\section{Experimental Setup}

In this section, we describe the datasets and the benchmarks used to train and evaluate our contribution.
\subsection{Datasets}
\subsubsection{Training Data}Similar to R2D2, our training dataset is a combination of synthetic pairs generated from random web images and Aachen dataset ~\cite{r2d2}. Note that we do not use any image from the HPatches evaluation dataset or our "UrbanScenes3D-Air" dataset during training.

\subsubsection{HPatches}
\label{sec:hpatches_description}
We use this dataset for evaluating our model on the task of feature matching. The dataset contains a total of 116 scenes, 57 of which relate to illumination changes and the remaining 59 to scene viewpoint variation. Each scene contains a set of six images and five ground truth homographies between the first and the rest of the five images. During the evaluation, we resize the image pairs to (300,300) in dimensions before matching to overcome the GPU OOM issues.
We also transform the ground-truth homographies according to the scaling done on the images. 

Further, we additionally rotate the rest of the five images in a given scene incrementally in the range of $0^\circ$ to $360^\circ$ to verify the feature matching robustness across the rotation angles. This results in performing matching on a total of $580$ image pairs at every angle and $13,920$ image pairs in total across all the angles. This rotation is an additional homography applied to the existing homographies between the image pairs.

\subsubsection{``UrbanScenes3D-Air'' Dataset for VPR}

While several Visual Place Recognition (VPR) datasets exist in the literature~\cite{zaffar2021vpr,warburg2020mapillary,garg2021your}\footnote{\cite{vallone2022danish} is a relevant dataset for this research but not yet released.}, the extent of viewpoint variation characteristics they offer is often either limited or tied to particular applications such as on-road driving~\cite{garg2018lost}. To complement the existing benchmark, we propose \textit{UrbanScenes3D-Air} dataset for UAV/drone based VPR. For this purpose, we imported the real-world Urbanscenes 3D dataset~\cite{liu2021urbanscene3d} into the AirSim environment~\cite{airsim2017fsr}. We then created a virtual drone equipped with downward-facing RGB camera and flew it over the buildings in the that environment to capture image sequences. To generate our UrbanScenes3D-Air dataset, the drone was flown in two trajectories as shown in the Figure~\ref{fig:airsim}. We pick the \textbf{yellow} trajectory for populating query database with 20 images taken at different locations. For this query image trajectory, we attach a single camera and fly the drone heading straight along the \textbf{yellow} path and capture the downward camera's views at regular intervals. Further, the reference database is populated with about 200 images from the \textbf{brown} trajectory for a given query image. While recording this brown trajectory we place 8 cameras at the bottom of the drone, each of which has a relative rotation in yaw. In Figure \ref{fig:airsim}, the 8 blue lines represent the setup of the cameras, each having a relative rotation with respect to drone's yaw axis. So, at a given location, we are able to capture 8 images which are relatively rotated in yaw. We do this to simulate other trajectories that might pass through the original trajectory at an angle but if done so would not have much trajectory overlap, thus this lets us test the same thing but with more image pairs in reference database at intersection. We formulate the problem of VPR here as retrieving the correct image match for a given query image from a set of images that have a relative rotation to the query. This is similar to the drone crossing a previous trajectory at an angle and re-localizing itself at the intersection of the previous and the current trajectory. The relative angle of rotation in our dataset varies from 0 to 90 and also includes opposite viewpoint images with 180 degree relative rotation. Figure~\ref{fig:airsim} depicts samples of query images which have a relative rotation with the the first image which is a sample of the reference database.
We further present our evaluation process of the VPR on our UrbanScenes-Air dataset in~\ref{subsec:vpr}

\begin{figure*}
    \centering
    \vspace*{0.1cm}
    \begin{tabular}{c c}
        \includegraphics[width=0.4\textwidth]{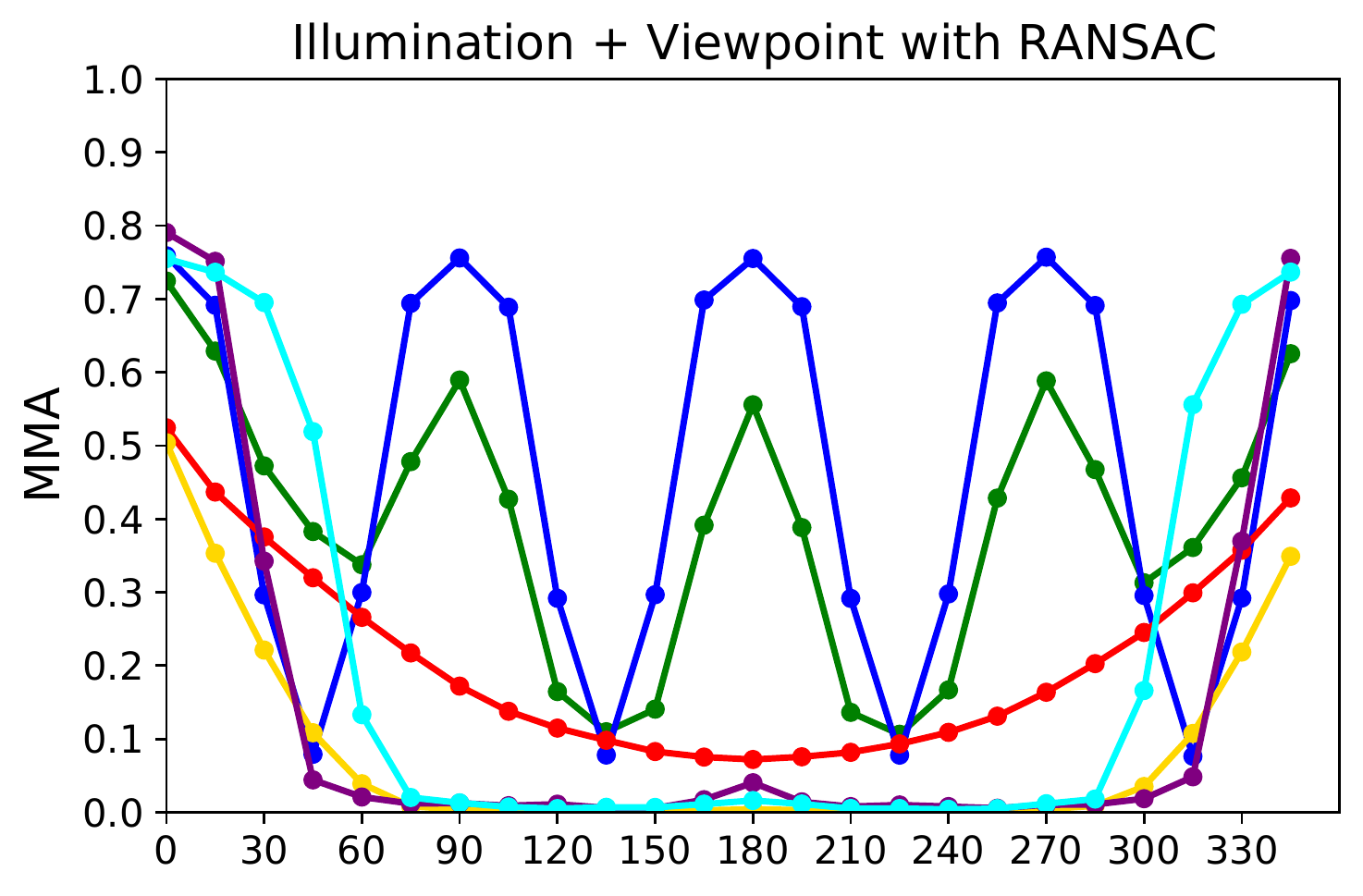} & \includegraphics[width=0.4\textwidth]{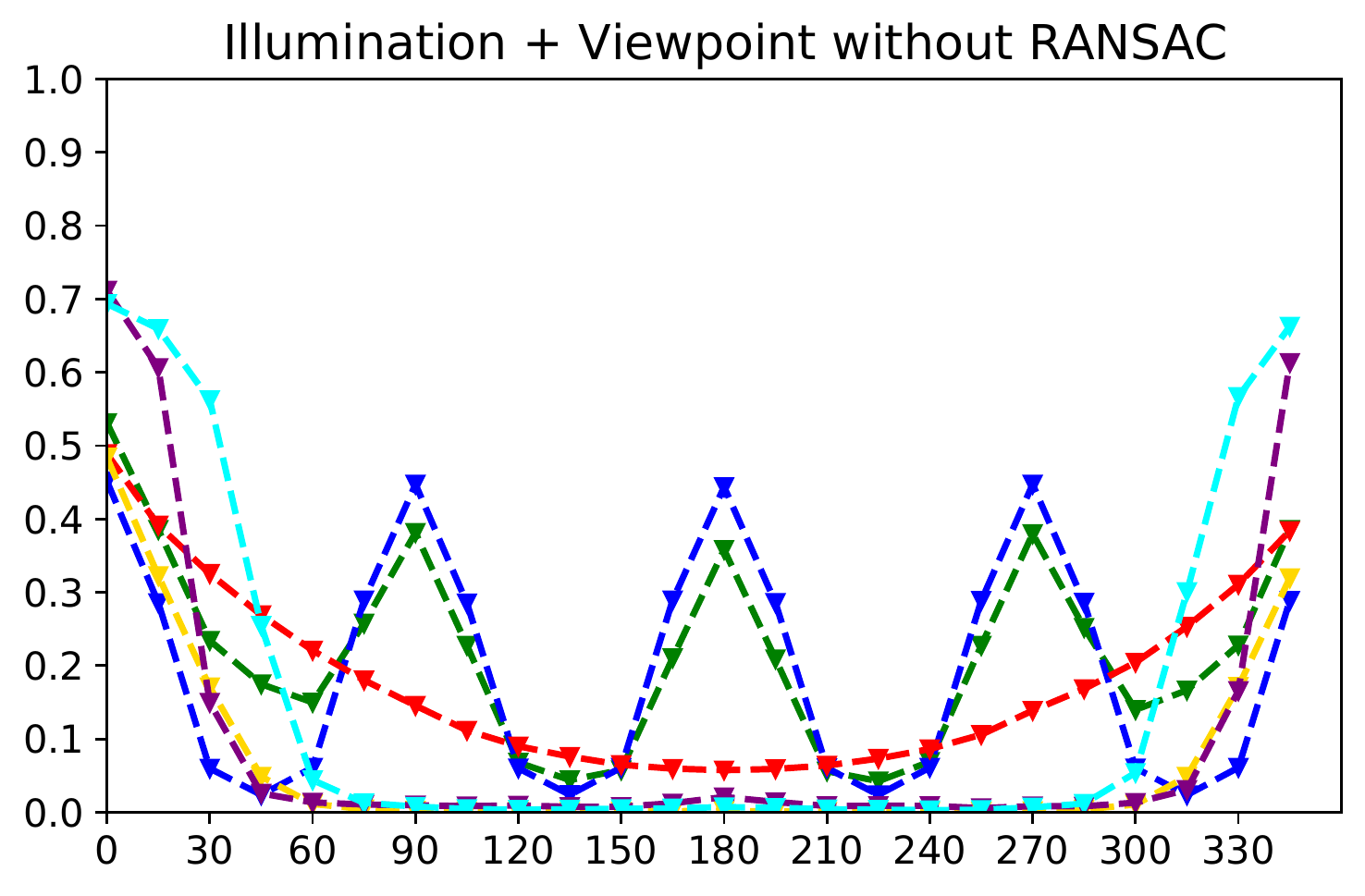} \\
        \includegraphics[width=0.4\textwidth]{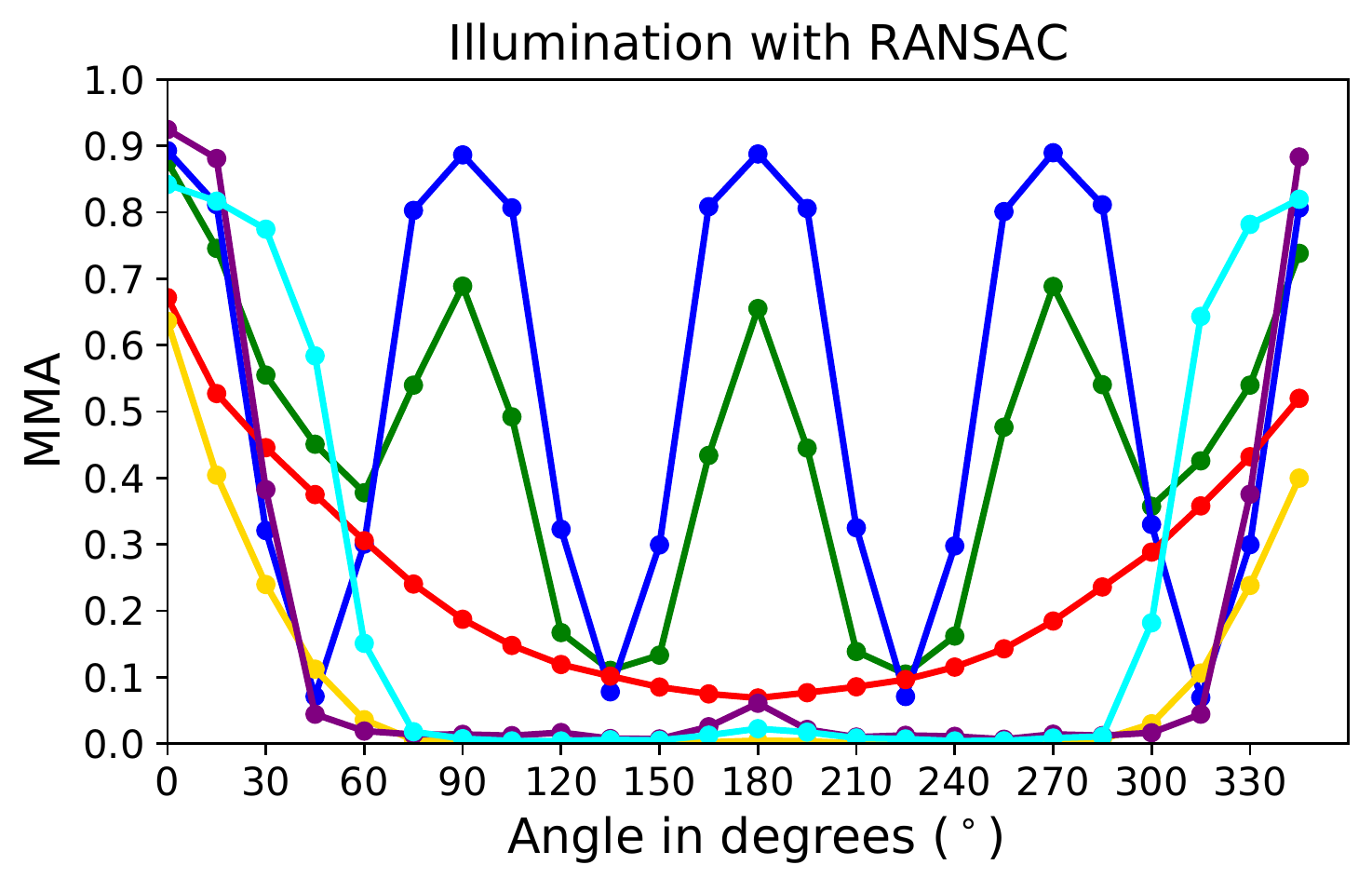} &
        \includegraphics[width=0.4\textwidth]{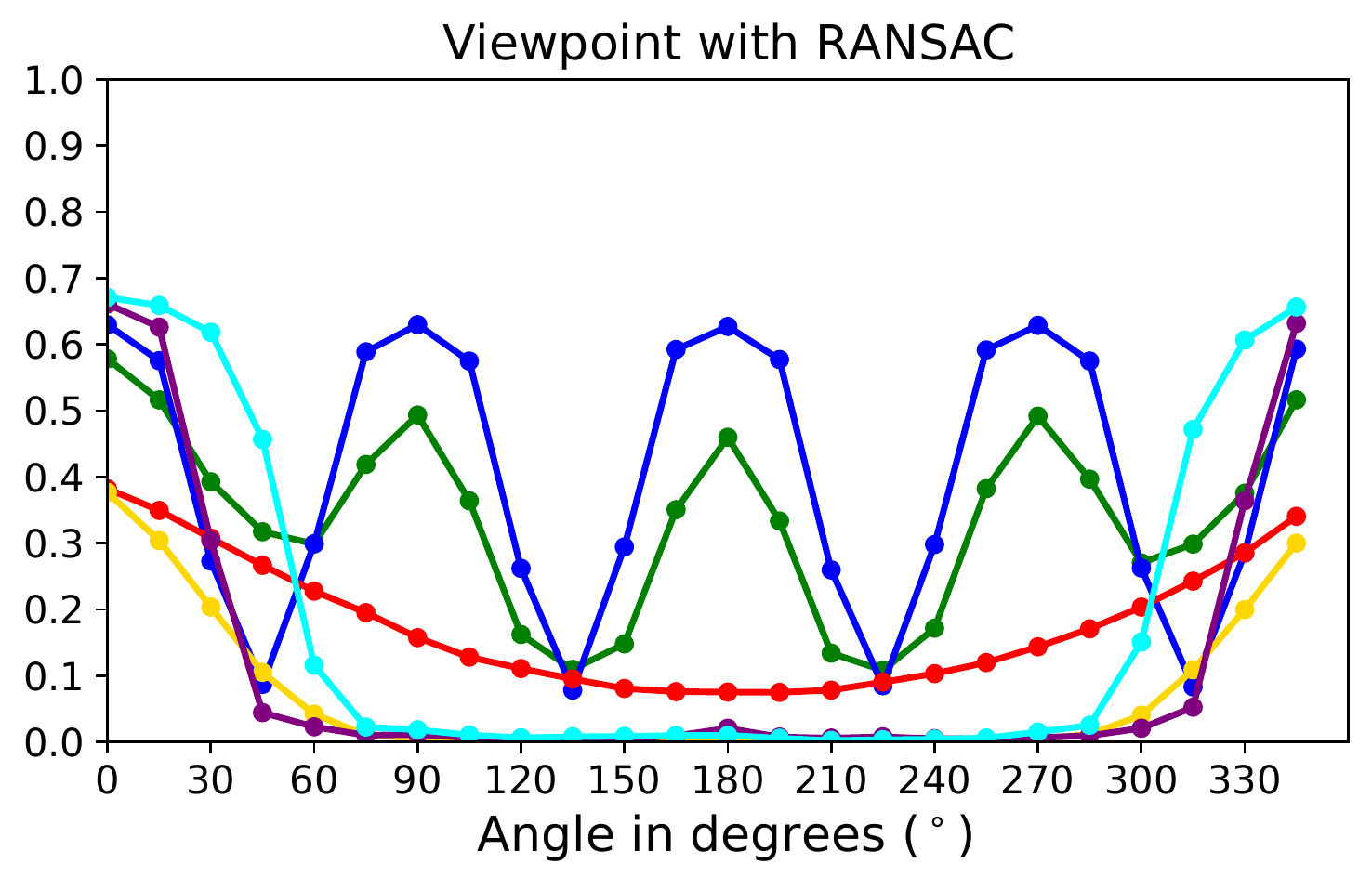} \\
        \includegraphics[width=0.4\textwidth]{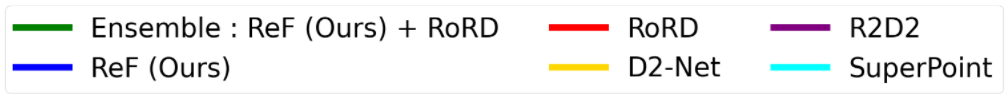}&
        \includegraphics[scale=0.6]{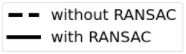}\\
    \end{tabular}
    \caption{The first row compares the performance of various methods on HPatches at different angles with and without RANSAC. The second row shows the individual performance of only illumination+rotation variations and viewpoint+rotation variations. ReF (Ours) outperforms all the other methods in comparison at certain angles in the neighbourhood of discrete C8 space like $75^\circ$, $90^\circ$ and $105^\circ$ when we apply RANSAC based filtering on the matches obtained from mutual nearest neighbours. Further, our proposed ensembling of ReF with RoRD improves the overall performance across rotation angles. All the scores are calculated for error threshold of 3 pixels.}
    \label{fig:mma_angle}
\end{figure*}
\subsection{Tasks and Evaluation}

\subsubsection{Local Feature Matching}

We evaluate the local feature matching on HPatches using Mean Matching Accuracy (MMA). A feature correspondence is considered to be correct if the back-projection of the keypoint using the ground truth homography matrix falls within a specific pixel threshold. We use pixel threshold of 3 for plots showing performance at various angles. Figure~\ref{fig:mma_pixel} shows performance averaged across all angles for different pixel thresholds.
We also compare the performance after filtering the correspondences with RANSAC since it is typically used to filter matches for tasks like VPR~\cite{hausler2021patch,rord} and 6-DoF pose estimation~\cite{sattler2018benchmarking}.

\subsubsection{Visual Place Recognition}
\label{subsec:vpr}
  We structured our \textit{UrbanScenes-Air} dataset so that the ideal match in the reference database has the same \verb|image_index| as the query image. After feature matching, our VPR pipeline returns the image in the reference database with the most inliers for a given query image. For evaluation, the predicted match is considered to be correct when its \verb|image_index| lies within a threshold of $\pm 2$ indices from the query image’s index. This is then repeated by changing the query database images to other angles like 15, 30, 45, 60, 75, 90, and 180. For each generated query database we evaluate and compare our ReF and the proposed ensemble of ReF+RoRD pipelines with other state-of-the-art feature matching pipelines. We plot the recall scores as a metric for VPR evaluation at various angles on our UrbanScenes3D-Air dataset in Figure~\ref{fig:airsim_plot}.

\section{RESULTS}
\label{sec:results}

In this section, we first present benchmark results for standalone local feature matching, then present both quantitative and qualitative results for VPR, which is finally followed by additional analyses of rotation robustness of E2-CNNs. 
\subsection{Local Feature Matching}
\label{sec:results_hpatches}

\subsubsection{Benchmarking across rotation angles}
The quantitative results for local feature matching on HPatches dataset is presented in Figure~\ref{fig:mma_angle}. This evaluation is performed on a rotated variant of HPatches as proposed in~\cite{rord} and described in~\ref{sec:hpatches_description} . Figure~\ref{fig:mma_angle} presents MMA scores for both with and without RANSAC (top row) along with split results on illumination and viewpoint alone (bottom row). It can be observed that ReF (blue) not just achieves peak performance at certain angles from its discrete C8 space but also in its immediate neighborhood, for example, $75^\circ$ and $105^\circ$ in the neighborhood of $90^\circ$. On the other hand, all methods including RoRD~\cite{rord} and SuperPoint~\cite{superpoint}, which are trained with heavy data augmentation for rotation robustness, exhibit meaningful MMA only under moderate rotation variations, for example, up to $\pm 60^\circ$. RoRD, however, performs the best among all existing data-augmented CNNs, and is thus used in our ensemble model (green) which improves the overall performance coverage across rotation angles with slight reduction of ReF's peak performance. This trend is further highlighted in Table~\ref{tab:airsim-table} which shows average MMA across all angles when using RANSAC.

\begin{figure}
    \centering
    \includegraphics[width=0.4\textwidth]{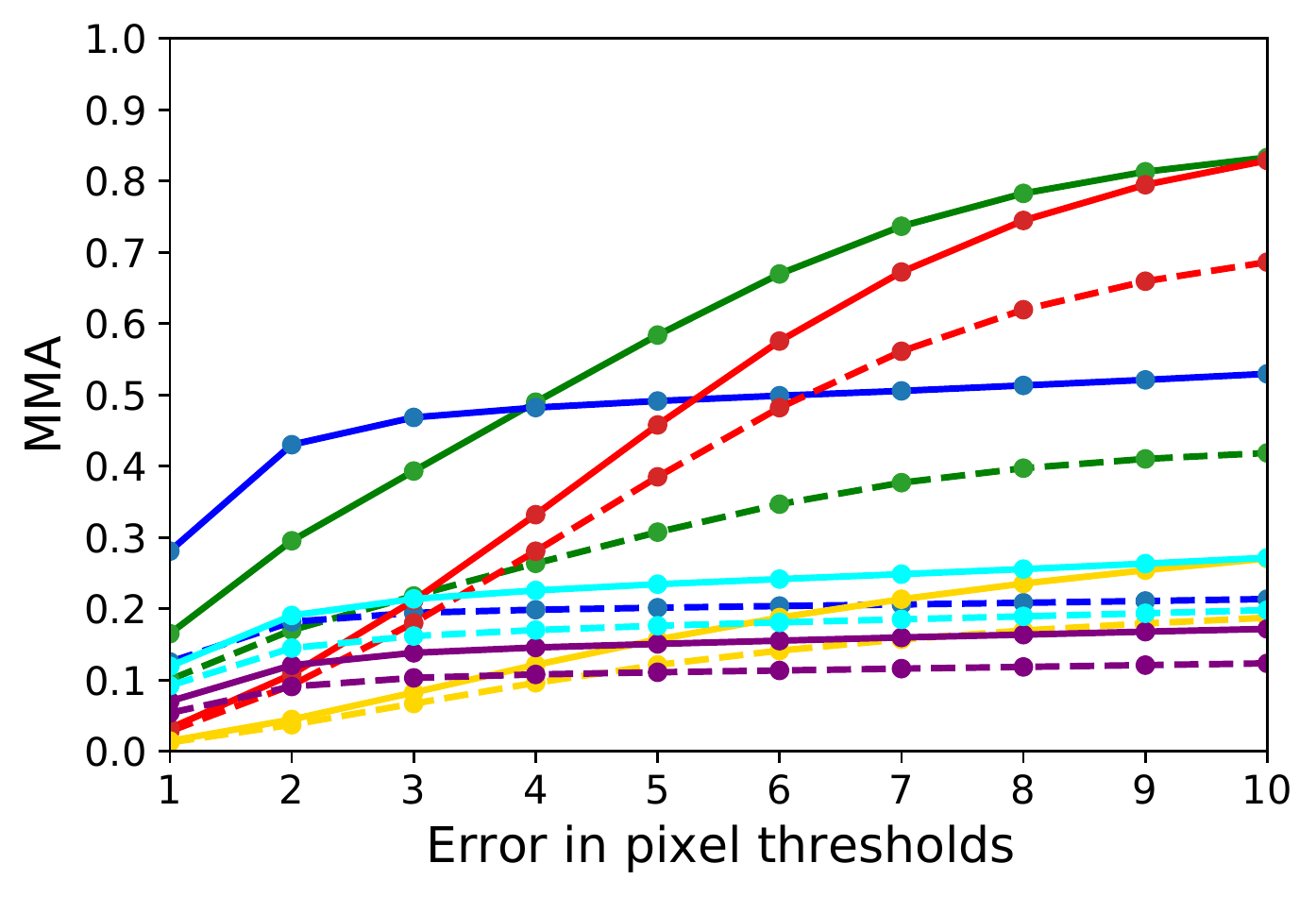}\\
    \includegraphics[width=0.4\textwidth]{images/L1.PNG}
    \includegraphics[scale=0.6]{images/L2.PNG}
    
    \caption{MMA values averaged across a range of angles in [$0^\circ$, $360^\circ$], at various pixel thresholds, for multiple methods. Ensemble outperforms other methods at higher pixel thresholds, where as ReF performs best at lower pixel thresholds.}
    \label{fig:mma_pixel}
\end{figure}

\begin{figure}
    \centering
    \includegraphics[width=0.4\textwidth]{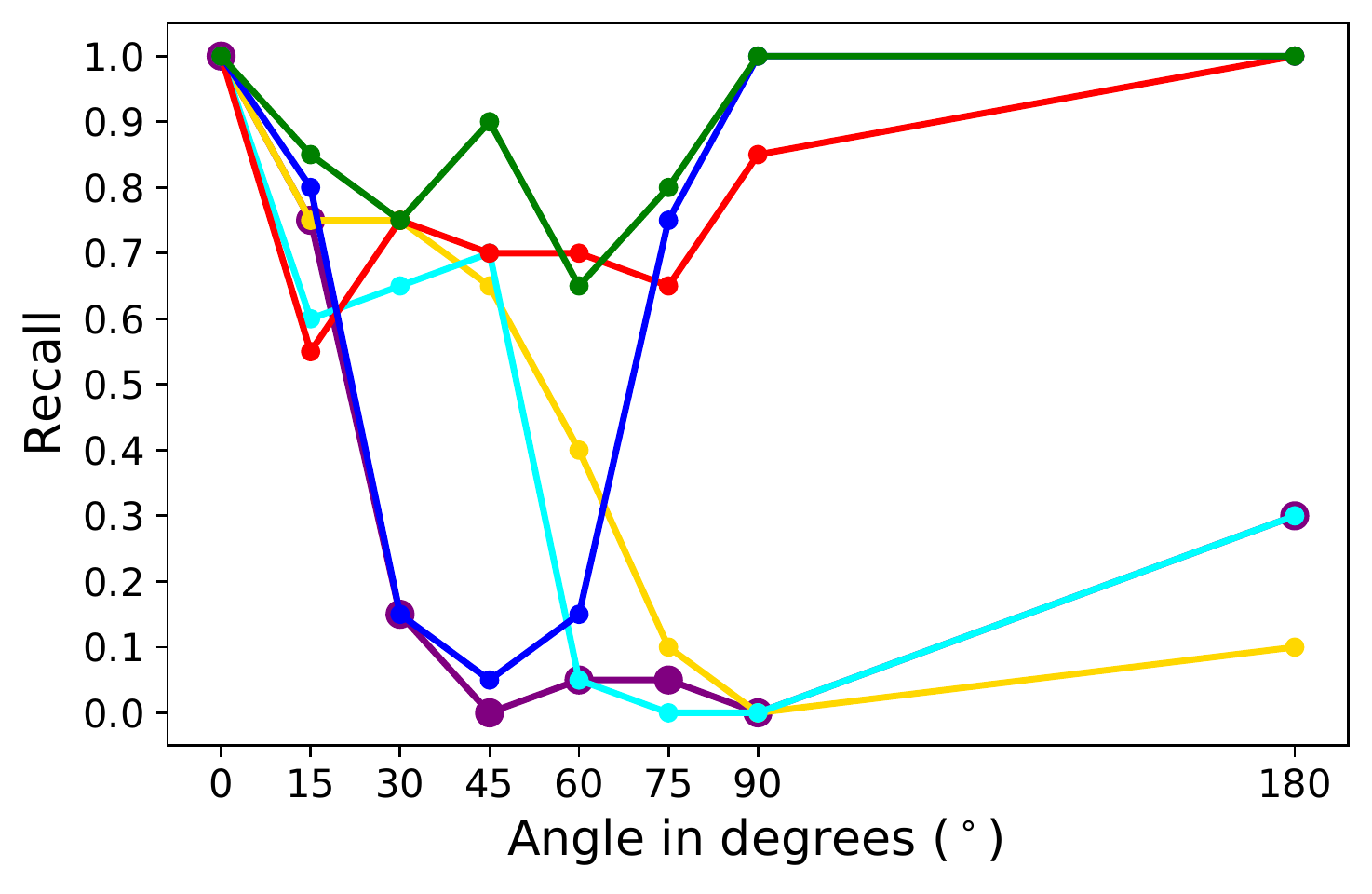}
    \includegraphics[width=0.4\textwidth]{images/L1.PNG}
    \caption{VPR on UrbanScenes3D-Air: The ensemble of ReF with RoRD outperforms all the other methods in comparison.}
    \label{fig:airsim_plot}
\end{figure}

\begin{table}
\centering
\vspace*{0.3cm}
 \begin{tabular}{||c | c c c||} 
 \hline
 Method & Recall(\%) & MMA(3px) & MMA(6px) \\ %
 \hline
 R2D2 & 28.75 & 0.138 & 0.156\\ 
 D2-Net & 46.88 & 0.0826 & 0.188\\
 SuperPoint & 41.25 & 0.214 & 0.242\\
 RoRD & 77.50 & 0.212 & 0.576\\
 ReF (\textit{ours}) & 61.25 & \textbf{0.469} & 0.499\\ 
 ReF (\textit{ours}) + RoRD & \textbf{86.88} & 0.393 & \textbf{0.669}  \\ %
 \hline
 \end{tabular}
 \caption{\label{tab:airsim-table}Recall and MMA scores averaged across rotation angles as evaluated on UrbanScenes3D-Air dataset for VPR and local feature matching on HPatches respectively. The average MMA values reported here were computed at pixel threshold 3 and 6 respectively in the last two columns, corresponding to the top left plot in Figure~\ref{fig:mma_angle} using RANSAC based filtering.}
\end{table}

\paragraph{With and Without RANSAC} In Figure~\ref{fig:mma_angle} top row, we compare results with and without RANSAC. It can be seen that while there is general improvement of MMA scores through RANSAC for all the methods, some methods benefit more than others, e.g. ReF (blue). This indicates that the feature learnt by ReF can lead to correct correspondences but with noisy outliers, which can then be successfully filtered by a robust outlier rejection method like RANSAC, thus validating ReF's practical use in tasks like VPR. Other methods do not show a similar behaviour since their initial correspondences set does not comprise sufficient correct matches. On the other hand, the ensemble method (green) is able to leverage the correct correspondences from both RoRD and ReF, where its performance in most cases is close to the better performing ingredient method (RoRD or ReF).

\paragraph{Illumination vs Viewpoint} In Figure~\ref{fig:mma_angle} bottom row, we compare performance split across illumination and viewpoint variations (with RANSAC). Here, the viewpoint variations refer to the perspective viewpoint change originally presented in the HPatches dataset. It can be observed that performance trends are maintained across the split results but the combined effect of perspectivity change on top of rotation variations reduces the absolute MMA. 

\subsubsection{Benchmarking at different pixel thresholds}
In the previous study, we considered the pixel threshold of 3.
Here, we observe the MMA scores averaged across all the rotation variations plotted against different pixel thresholds in Figure~\ref{fig:mma_pixel}. The ensemble method (solid green) outperforms all the other methods across the board except the tighter thresholds where ReF (solid blue) performs the best. It can also be observed that use of RANSAC (solid) changes the performance trends as compared to without RANSAC (dashed). This justifies the need for RANSAC to correctly select the right set of correspondences from an initial set of matches which is far less noisy than that obtained from the other state-of-the-art methods.

\subsection{VPR on UrbanScenes3D-Air}
The performance of various feature matching pipelines on our \textit{UrbanScenes3D-Air} dataset for the task of VPR can be seen in Figure~\ref{fig:airsim_plot}. As the relative angle of rotation increases from 0 to 90 degrees, ReF(Ours) performs significantly well at discrete higher rotations when compared to the other state-of-the-art methods. Further, the ensemble of our ReF with RoRD improves the recall scores at specific angles which are not handled by ReF and thus outperforms all the other methods in comparison. The same trend is also summarized in Table~\ref{tab:airsim-table}. The qualitative matches shown in Figure~\ref{fig:vpr_qualitative} highlight the superior performance of ReF model under large rotation variations with rich and accurate correspondences where methods like R2D2~\cite{r2d2} and SuperPoint~\cite{superpoint} fail.

\begin{figure}
    \centering
    \vspace*{0.1cm}
    \includegraphics[width=0.47\textwidth]{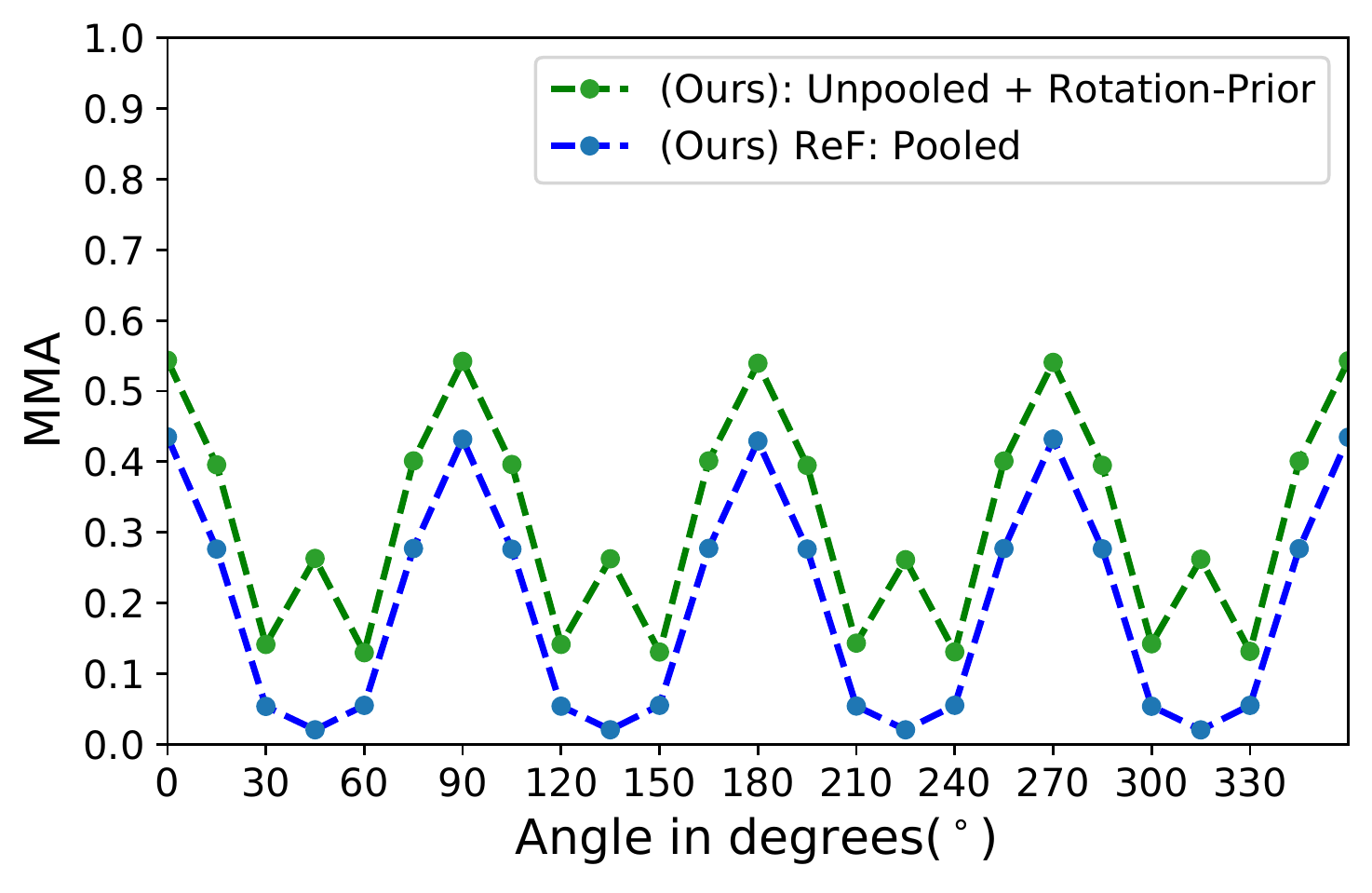}
    \caption{We can clearly see that our unpooled version performs better than the group pooled version when a rotation prior is given at angles where the group pooled version fails like $45^\circ$, $135^\circ$, etc.}
    \label{fig:mma_rotation}
\end{figure}

\subsection{Ablation Studies}

\subsubsection{Performance with Rotation Priors}
The E2-CNN layers are equivariant under discrete rotations and we are able to extract feature map specific to a specific angle. So instead of group pooling the feature fields obtained from E2-CNN in our network, we directly use the unpooled features. 
Here, we evaluate the performance in a scenario where we have a rotation prior available. That is, we know the relative orientation between the two images for matching, which in some practical use cases could be known, for example, back and forth last mile deliveries on repeated routes. Figure~\ref{fig:mma_rotation} shows the MMA results for this setting. It can be seen that the features learned by C8 space are better highlighted in this scenario where there is a performance jump at $45^\circ$ as well.

\begin{figure}
    \centering
    \vspace*{0.1cm}
    \includegraphics[width=0.4\textwidth]{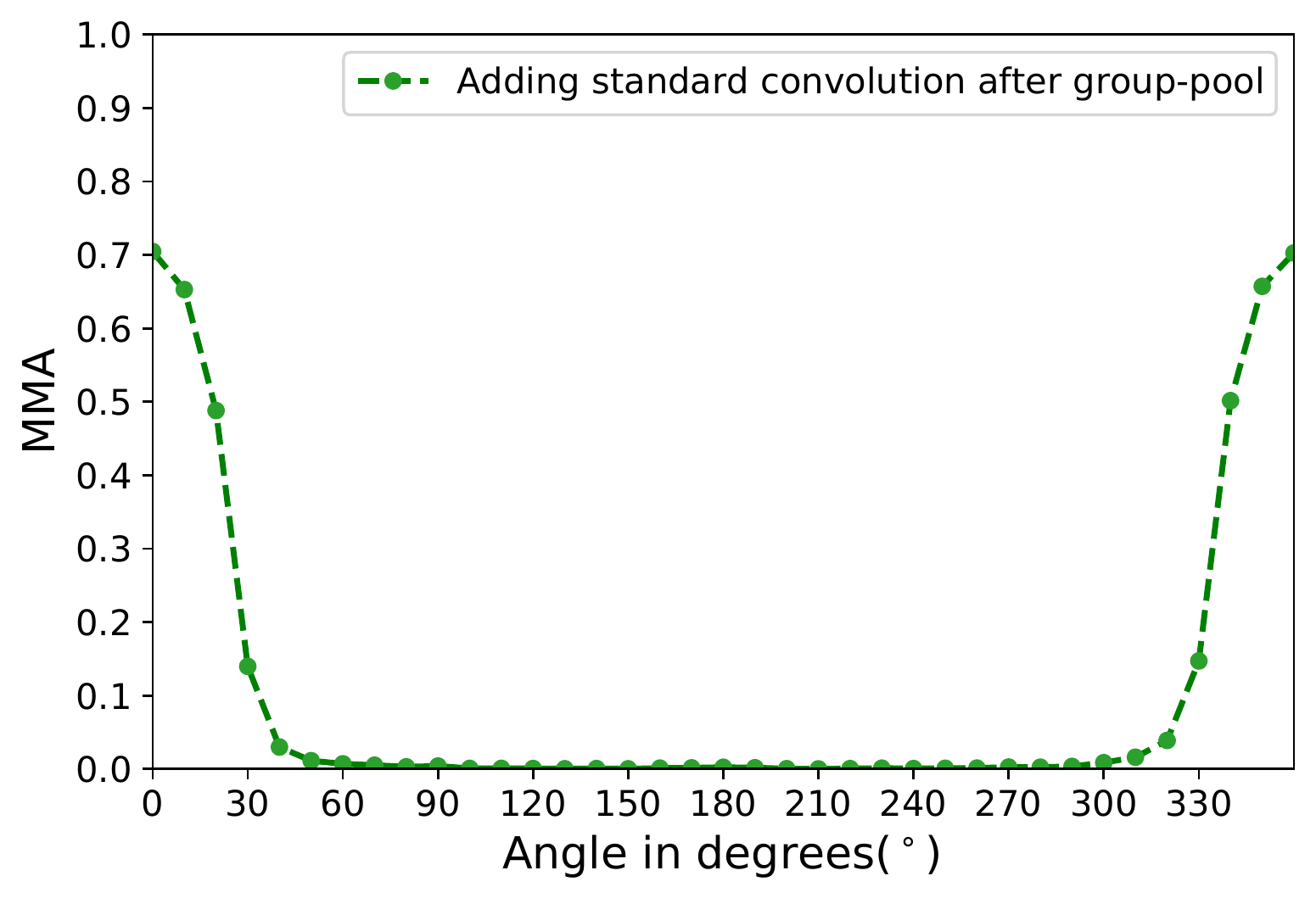}
    \caption{Equivariance is lost when standard convolutions are added after the group pooling layer.}
    \label{fig:mma_cnn}
\end{figure}

\begin{figure}
    \centering
    \includegraphics[width=0.4\textwidth]{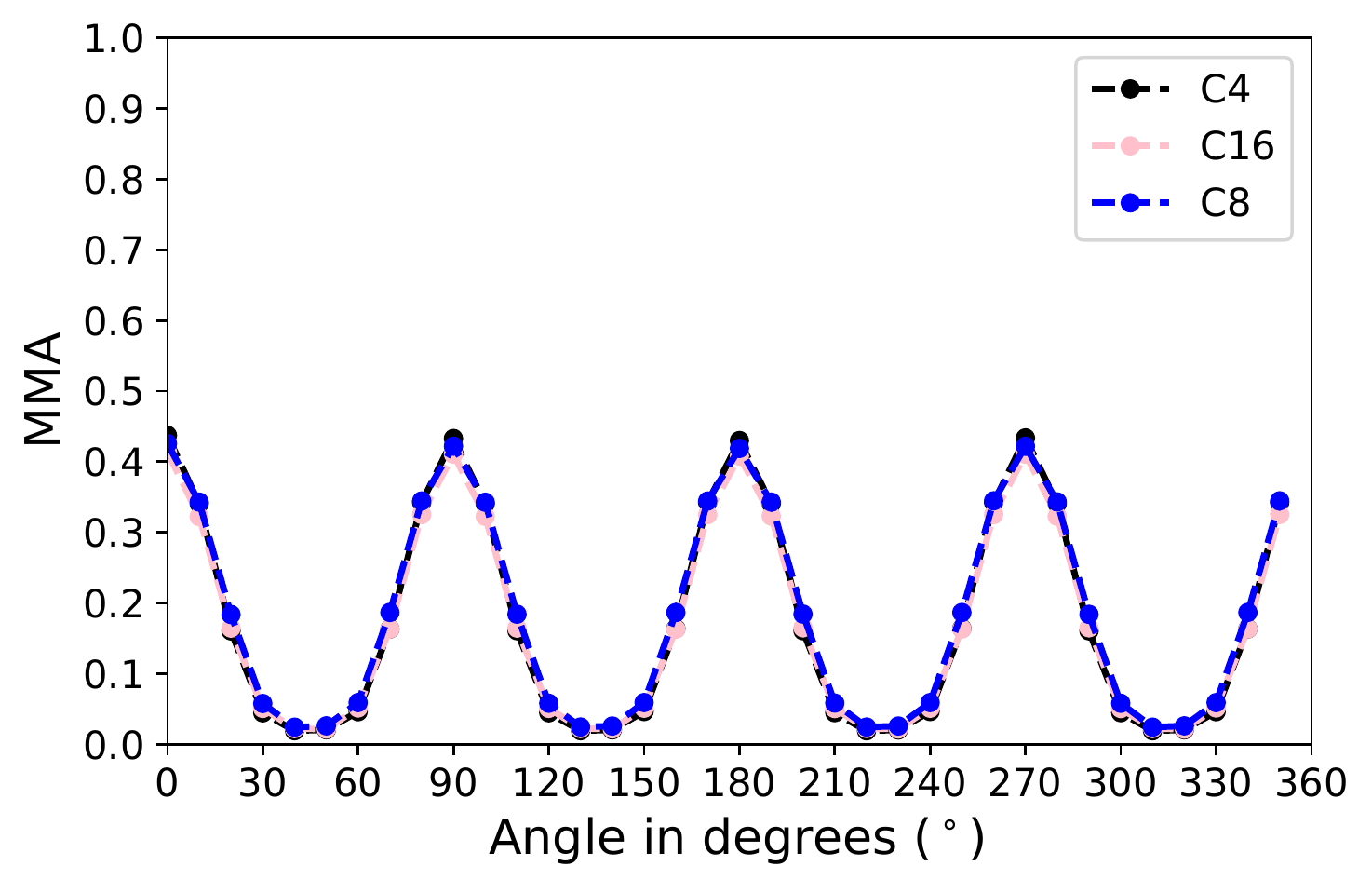}
    \caption{Performance comparison of various cyclic groups $C_N$ for $N \in \{4,8,16\}$.}
    \label{fig:cyclic_comparison}
\end{figure}

\subsubsection{Effect of mixing Standard CNN with E2-CNN} 
In this section we evaluate the impact of adding standard CNN layers after group pooling layer for the descriptors.
A lot of precious work on classification~\cite{e2cnn,scnn} have shown various model ablations which combine different types of layers. 
For this experiment we add 3 standard CNN layers with filter size 2 with stride increment of 4,8 and 16 after group pooling layer.
Figure~\ref{fig:mma_cnn} shows that the equivariance property is completely lost, and the performance is flattened out across the angles, whereas in Figure~\ref{fig:mma_angle} the performance of ReF peaks at specific angles.
 So, directly using the group pooled features as descriptors is necessary for local feature training to work. Although our ReF-network has 2 standard CNN layers to the end, those are only for classification and loss computation, and do not operate on the predicted descriptors after group pooling, thus preserving the invariance. 
 
\subsubsection{Effect of using different cyclic groups}
In Figure~\ref{fig:cyclic_comparison}, we compare the performance of various cyclic groups $C_N$ for $N \in \{4,8,16\}$. It can be observed that performance does not change noticeably across different values. Our choice of $C_8$ was based on a balance between performance and compute given that in the unpooled settings (that is with a given rotation prior, see Figure~\ref{fig:mma_rotation}), one could still achieve performance response on angles like $45^\circ$ which is not possible by definition for $C_4$ space.

\section{CONCLUSIONS}
In this work we present a novel architecture which leverages a steerable CNN for local feature matching between images under rotation variations. We further extend the capability of our model by introducing an ensemble approach combining ReF + RoRD, which combines the benefits of both models and improves overall performance across all possible rotations. We show that this ensemble outperforms other methods on the UrbanScenes3D-Air dataset, and on HPatches at higher pixel thresholds.

We explore the possibility for using steerable CNN for feature matching. This work can further be extended by studying the effects of different network architecture nuances like residual connections and combining steerable CNN on different spaces, e.g. C16, C8, C4, in the same network. Our study shows that by naively combining steerable CNN with standard CNN the equivariance property is lost. But more rigorous study on this front could lead to more novel architectures which help in feature matching under various rotations and extreme viewpoint changes.

\bibliographystyle{IEEEtran}
\bibliography{IEEEexample}

\begin{thebibliography}{10}
\providecommand{\url}[1]{#1}
\csname url@rmstyle\endcsname
\providecommand{\newblock}{\relax}
\providecommand{\bibinfo}[2]{#2}
\providecommand\BIBentrySTDinterwordspacing{\spaceskip=0pt\relax}
\providecommand\BIBentryALTinterwordstretchfactor{4}
\providecommand\BIBentryALTinterwordspacing{\spaceskip=\fontdimen2\font plus
\BIBentryALTinterwordstretchfactor\fontdimen3\font minus
  \fontdimen4\font\relax}
\providecommand\BIBforeignlanguage[2]{{%
\expandafter\ifx\csname l@#1\endcsname\relax
\typeout{** WARNING: IEEEtran.bst: No hyphenation pattern has been}%
\typeout{** loaded for the language `#1'. Using the pattern for}%
\typeout{** the default language instead.}%
\else
\language=\csname l@#1\endcsname
\fi
#2}}

\bibitem{cummins2011appearance}
M.~Cummins and P.~Newman, ``Appearance-only slam at large scale with fab-map
  2.0,'' \emph{The International Journal of Robotics Research}, vol.~30, no.~9,
  pp. 1100--1123, 2011.

\bibitem{hausler2021patch}
S.~Hausler, S.~Garg, M.~Xu, M.~Milford, and T.~Fischer, ``Patch-netvlad:
  Multi-scale fusion of locally-global descriptors for place recognition,'' in
  \emph{Proceedings of the IEEE/CVF Conference on Computer Vision and Pattern
  Recognition}, 2021, pp. 14\,141--14\,152.

\bibitem{garg2021your}
S.~Garg, T.~Fischer, and M.~Milford, ``Where is your place, visual place
  recognition?'' in \emph{Proceedings of the Thirtieth International Joint
  Conference on Artificial Intelligence (IJCAI-21)}.\hskip 1em plus 0.5em minus
  0.4em\relax International Joint Conferences on Artificial Intelligence, 2021,
  pp. 4416--4425.

\bibitem{image_stitching}
\BIBentryALTinterwordspacing
T.~Liao and N.~Li, ``Single-perspective warps in natural image stitching,''
  \emph{IEEE Transactions on Image Processing}, vol.~29, p. 724–735, 2020.
  [Online]. Available: \url{http://dx.doi.org/10.1109/TIP.2019.2934344}
\BIBentrySTDinterwordspacing

\bibitem{sattler2018benchmarking}
T.~Sattler, W.~Maddern, C.~Toft, A.~Torii, L.~Hammarstrand, E.~Stenborg,
  D.~Safari, M.~Okutomi, M.~Pollefeys, J.~Sivic, \emph{et~al.}, ``Benchmarking
  6dof outdoor visual localization in changing conditions,'' in
  \emph{Proceedings of the IEEE conference on computer vision and pattern
  recognition}, 2018, pp. 8601--8610.

\bibitem{superglue}
\BIBentryALTinterwordspacing
P.-E. Sarlin, D.~DeTone, T.~Malisiewicz, and A.~Rabinovich, ``{SuperGlue}:
  Learning feature matching with graph neural networks,'' in \emph{CVPR}, 2020.
  [Online]. Available: \url{https://arxiv.org/abs/1911.11763}
\BIBentrySTDinterwordspacing

\bibitem{rord}
U.~S. Parihar, A.~Gujarathi, K.~Mehta, S.~Tourani, S.~Garg, M.~Milford, and
  K.~M. Krishna, ``Rord: Rotation-robust descriptors and orthographic views for
  local feature matching,'' 2021.

\bibitem{d2net}
M.~Dusmanu, I.~Rocco, T.~Pajdla, M.~Pollefeys, J.~Sivic, A.~Torii, and
  T.~Sattler, ``{D2-Net: A Trainable CNN for Joint Detection and Description of
  Local Features},'' in \emph{Proceedings of the 2019 IEEE/CVF Conference on
  Computer Vision and Pattern Recognition}, 2019.

\bibitem{sift}
D.~Lowe, ``Distinctive image features from scale-invariant keypoints,''
  \emph{International Journal of Computer Vision}, vol.~60, pp. 91--, 11 2004.

\bibitem{orb}
E.~Rublee, V.~Rabaud, K.~Konolige, and G.~Bradski, ``Orb: an efficient
  alternative to sift or surf,'' 11 2011, pp. 2564--2571.

\bibitem{superpoint}
\BIBentryALTinterwordspacing
D.~DeTone, T.~Malisiewicz, and A.~Rabinovich, ``Superpoint: Self-supervised
  interest point detection and description,'' \emph{CoRR}, vol. abs/1712.07629,
  2017. [Online]. Available: \url{http://arxiv.org/abs/1712.07629}
\BIBentrySTDinterwordspacing

\bibitem{scnn}
\BIBentryALTinterwordspacing
T.~S. Cohen and M.~Welling, ``Steerable cnns,'' \emph{CoRR}, vol.
  abs/1612.08498, 2016. [Online]. Available:
  \url{http://arxiv.org/abs/1612.08498}
\BIBentrySTDinterwordspacing

\bibitem{e2cnn}
M.~Weiler and G.~Cesa, ``{General E(2)-Equivariant Steerable CNNs},'' in
  \emph{Conference on Neural Information Processing Systems (NeurIPS)}, 2019.

\bibitem{r2d2}
J.~Revaud, P.~Weinzaepfel, C.~D. Souza, N.~Pion, G.~Csurka, Y.~Cabon, and
  M.~Humenberger, ``R2d2: Repeatable and reliable detector and descriptor,''
  2019.

\bibitem{LHF}
T.~Ahonen, J.~Matas, C.~He, and M.~Pietikäinen, ``Rotation invariant image
  description with local binary pattern histogram fourier features,'' 06 2009,
  pp. 61--70.

\bibitem{lbp}
T.~{Ojala}, M.~{Pietikainen}, and T.~{Maenpaa}, ``Multiresolution gray-scale
  and rotation invariant texture classification with local binary patterns,''
  \emph{IEEE Transactions on Pattern Analysis and Machine Intelligence},
  vol.~24, no.~7, pp. 971--987, 2002.

\bibitem{lribp}
Y.~{Duan}, J.~{Lu}, J.~{Feng}, and J.~{Zhou}, ``Learning rotation-invariant
  local binary descriptor,'' \emph{IEEE Transactions on Image Processing},
  vol.~26, no.~8, pp. 3636--3651, 2017.

\bibitem{keetha2021hierarchical}
N.~V. Keetha, M.~Milford, and S.~Garg, ``A hierarchical dual model of
  environment-and place-specific utility for visual place recognition,''
  \emph{IEEE Robotics and Automation Letters}, vol.~6, no.~4, pp. 6969--6976,
  2021.

\bibitem{wang2022transvpr}
R.~Wang, Y.~Shen, W.~Zuo, S.~Zhou, and N.~Zhen, ``Transvpr: Transformer-based
  place recognition with multi-level attention aggregation,'' 2022.

\bibitem{scnn_pdo}
\BIBentryALTinterwordspacing
E.~Jenner and M.~Weiler, ``Steerable partial differential operators for
  equivariant neural networks,'' \emph{CoRR}, vol. abs/2106.10163, 2021.
  [Online]. Available: \url{https://arxiv.org/abs/2106.10163}
\BIBentrySTDinterwordspacing

\bibitem{mri_roteq}
\BIBentryALTinterwordspacing
P.~M{\"{u}}ller, V.~Golkov, V.~Tomassini, and D.~Cremers,
  ``Rotation-equivariant deep learning for diffusion {MRI},'' \emph{CoRR}, vol.
  abs/2102.06942, 2021. [Online]. Available:
  \url{https://arxiv.org/abs/2102.06942}
\BIBentrySTDinterwordspacing

\bibitem{redet}
J.~Han, J.~Ding, N.~Xue, and G.-S. Xia, ``Redet: A rotation-equivariant
  detector for aerial object detection,'' 2021.

\bibitem{aircraft_detection}
\BIBentryALTinterwordspacing
X.~Chen, J.~Liu, F.~Xu, Z.~Xie, Y.~Zuo, and L.~Cao, ``A novel method of
  aircraft detection under complex background based on circular intensity
  filter and rotation invariant feature,'' \emph{Sensors}, vol.~22, no.~1,
  2022. [Online]. Available: \url{https://www.mdpi.com/1424-8220/22/1/319}
\BIBentrySTDinterwordspacing

\bibitem{rl_geq}
A.~K. Mondal, P.~Nair, and K.~Siddiqi, ``Group equivariant deep reinforcement
  learning,'' \emph{arXiv preprint arXiv:2007.03437}, 2020.

\bibitem{lin-rep-fg-book}
J.-P. Serre, \emph{Linear Representations of Finite Groups}, 1st~ed.\hskip 1em
  plus 0.5em minus 0.4em\relax Springer, 1977.

\bibitem{aploss}
K.~He, Y.~Lu, and S.~Sclaroff, ``Local descriptors optimized for average
  precision,'' 2018.

\bibitem{zaffar2021vpr}
M.~Zaffar, S.~Garg, M.~Milford, J.~Kooij, D.~Flynn, K.~McDonald-Maier, and
  S.~Ehsan, ``Vpr-bench: An open-source visual place recognition evaluation
  framework with quantifiable viewpoint and appearance change,''
  \emph{International Journal of Computer Vision}, vol. 129, no.~7, pp.
  2136--2174, 2021.

\bibitem{warburg2020mapillary}
F.~Warburg, S.~Hauberg, M.~Lopez-Antequera, P.~Gargallo, Y.~Kuang, and
  J.~Civera, ``Mapillary street-level sequences: A dataset for lifelong place
  recognition,'' in \emph{Proceedings of the IEEE/CVF conference on computer
  vision and pattern recognition}, 2020, pp. 2626--2635.

\bibitem{vallone2022danish}
A.~Vallone, F.~Warburg, H.~Hansen, S.~Hauberg, and J.~Civera, ``Danish airs and
  grounds: A dataset for aerial-to-street-level place recognition and
  localization,'' \emph{arXiv preprint arXiv:2202.01821}, 2022.

\bibitem{garg2018lost}
S.~Garg, N.~Suenderhauf, and M.~Milford, ``Lost? appearance-invariant place
  recognition for opposite viewpoints using visual semantics,''
  \emph{Proceedings of Robotics: Science and Systems XIV}, 2018.

\bibitem{liu2021urbanscene3d}
Y.~Liu, F.~Xue, and H.~Huang, ``Urbanscene3d: A large scale urban scene dataset
  and simulator,'' 2021.

\bibitem{airsim2017fsr}
\BIBentryALTinterwordspacing
S.~Shah, D.~Dey, C.~Lovett, and A.~Kapoor, ``Airsim: High-fidelity visual and
  physical simulation for autonomous vehicles,'' in \emph{Field and Service
  Robotics}, 2017. [Online]. Available: \url{https://arxiv.org/abs/1705.05065}
\BIBentrySTDinterwordspacing

\end{thebibliography}

\end{document}